\pdfoutput=1
\documentclass[11pt]{article}

\usepackage[preprint]{acl}

\usepackage[T1]{fontenc}
\usepackage[utf8]{inputenc}

\usepackage{microtype}

\usepackage{inconsolata}

\usepackage{graphicx}

\usepackage{amsmath}
\usepackage{multirow}
\usepackage{booktabs}
\usepackage{colortbl}
\usepackage{amssymb}
\usepackage{bbding}
\usepackage{pifont}
\usepackage{wasysym}
\usepackage{utfsym}
\usepackage{fontawesome}
\usepackage{siunitx}
\usepackage{soul}
\usepackage{pifont}
\usepackage{xcolor}
\usepackage{times}
\usepackage{latexsym}
\usepackage{amsmath}
\usepackage{amssymb}
\usepackage[ruled,linesnumbered,vlined]{algorithm2e}
\SetAlCapFnt{\small}         % caption 标题字体
\SetAlCapNameFnt{\small}     % "Algorithm" 前缀字体
\definecolor{blue}{HTML}{99CCFF}

\definecolor{red}{HTML}{F1C0BE}

\definecolor{green-black}{HTML}{16AB00}
\newcommand{\gtext}[1]{\textcolor{green-black}{#1}}
\definecolor{red1}{HTML}{FF0000}
\newcommand{\rtext}[1]{\textcolor{red1}{#1}}

\title{Mitigating Spurious Correlations Between Question and Answer via Chain-of-Thought Correctness Perception Distillation}

\author{
 \textbf{Hongyan Xie\textsuperscript{1}},
 \textbf{Yitong Yao\textsuperscript{2}},
 \textbf{Yikun Ban\textsuperscript{1}},
  \textbf{Zixuan Huang\textsuperscript{1}},
 \textbf{Deqing Wang\textsuperscript{1}},
\\
 \textbf{Zhenhe Wu\textsuperscript{1}},
\textbf{Haoxiang Su\textsuperscript{2}},
 \textbf{Chao Wang\textsuperscript{2}},
 \textbf{Shuangyong Song\textsuperscript{2}},
\\
\\
 \textsuperscript{1}  School of Computer, Beihang University, \\
 \textsuperscript{2} Institute of Artificial Intelligence (TeleAI), China Telecom,
\\
 \small{
   \href{mailto:xiehongyan@buaa.edu.cn}{xiehongyan@buaa.edu.cn}
 }
}

\begin{document}
\maketitle
\begin{abstract}
Large language models (LLMs) excel at reasoning tasks but are expensive to deploy. Thus small language models (SLMs) are fine-tuned on CoT data generated by LLMs to copy LLMs' abilities. However, these CoT data may include noisy rationales that either fail to substantiate the answers or contribute no additional information to support answer prediction, which leads SLMs to capture spurious correlations between questions and answers and compromise the quality of reasoning.
In this work, we propose Chain-of-Thought Correctness Perception Distillation (CoPeD), which aims to improve the reasoning quality of the student model from the perspectives of task setting and data utilization. Firstly, we introduce a correctness-aware task setting that encourages the student model to predict answers based on correct rationales and revise them when they are incorrect. This setting improves the faithfulness of reasoning and allows the model to learn from its mistakes. Then, we propose a Correctness-Aware Weighted loss, which dynamically adjusts the contribution of each training instance based on the combined loss of the rationale and the answer. This strategy encourages the model to focus more on samples where the rationale offers stronger support for the correct answer. Experiments have shown that CoPeD is effective on both in-distribution (IND) and out-of-distribution (OOD) benchmark reasoning datasets\footnote{We will release our code and data upon publication to facilitate reproducibility.}.

\end{abstract}

% This encourages the student model to rely on valid reasoning paths for answer prediction and learn from mistakes, thereby enhancing the faithfulness and soundness of the generated rationales.

\section{Introduction}

Through progressive scaling of model architectures and training datasets, LLMs have demonstrated exceptional CoT reasoning capabilities in complex NLP tasks. As evidenced by recent studies \cite{brown2020language,hoffmann2022training,chowdhery2023palm,openai2023gpt, chen2023mcc}, the CoT paradigm enables multi-step logical reasoning through explicit intermediate derivations. While this paradigm facilitates complex problem-solving, it also introduces significant computational costs. These costs pose practical challenges for real-world deployment \cite{ai_flow}. 
For example, GPT-3 \cite{brown2020language} has 175 billion parameters. Its inference requires substantial computation, making deployment costly.

\begin{figure}[tp]
  \includegraphics[width=0.90\columnwidth]{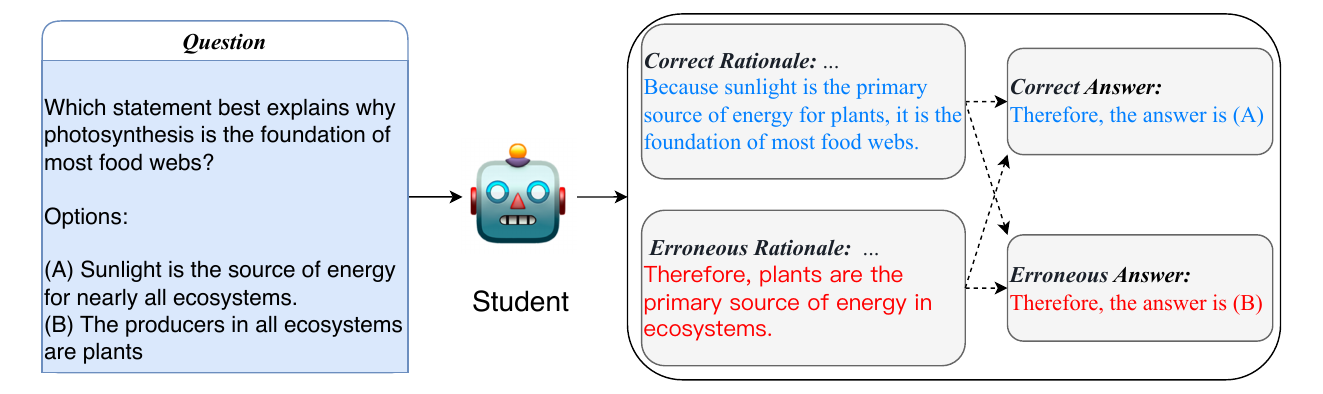}
      \vspace{-5pt}
	\centering
  \caption{During training, the student model may capture spurious correlations between the question and the answer. As a result, during inference, the rationale could be correct while the answer is erroneous, or the answer could be correct while the rationale is erroneous.}
  \label{fig:bg}
\end{figure}

Therefore, the current research \cite{west2022symbolic, magister2023teaching, ho2023large, fu2023specializing, chen2023disco, zhou2024teaching, li2024mode, chenglin2024mixed, wadhwa2024investigating, lee2024mentor} on knowledge distillation  aims to transfer the powerful reasoning ability of LLMs to SLMs. The standard process of this procedure consists of two stages: First, the LLM serves as a teacher to generate rationales for each sample. Then, these rationales are used to perform supervised fine-tuning on the SLM.
Although this paradigm improves the reasoning capabilities of SLMs on specific tasks, it commonly assumes that the generated rationale is reliable as long as the predicted answer is correct. However, this assumption does not always hold, as in many cases the rationale neither introduces new information beyond what is provided in the input nor effectively justifies the answer \cite{wang2023scott}. 
Therefore, during training, SLMs may capture spurious correlations between questions and answers, leading to two main issues, as shown in  Figure \ref{fig:bg}. First, the model may overlook the causal logical relationship between rationales and answers. As a result, the generated rationales may be inconsistent with the predicted answers \cite{wang2023scott, feng2024teaching}. 
Second, such spurious correlations can degrade the quality of rationale generation during reasoning \cite{dai-etal-2024-improve-students}. In particular, an error in intermediate steps may lead to error propagation in subsequent reasoning.

To address the above issues, we propose the \textbf{C}hain-\textbf{o}f-Thought Correctness \textbf{Pe}rception \textbf{D}istillation (CoPeD). Specifically, we begin by prompting the teacher model (LLM) to generate both correct and erroneous rationales. Following previous work \cite{dai2024beyond,dai-etal-2024-improve-students},  we assume 
that \textit{If the LLM’s predicted answer is correct, the rationale is assumed to be correct; otherwise, the rationale is considered erroneous.}
Building on this, we introduce a \textbf{correctness-aware task setting}. In this setting, the student model predicts the answer based on the question and rationale when the rationale is correct. Otherwise, it generates a revised rationale when the original rationale is erroneous. This task design encourages the student model to rely on valid reasoning paths for answer prediction, thereby mitigating spurious correlations between questions and answers.
Since the assumption regarding rationale correctness may not always hold \cite{wang2023scott}, we further propose a \textbf{correctness\footnote{Here, “correctness” refers to whether the rationale provides effective support for the ground-truth answer.}-aware confidence-weighted loss}. This loss dynamically adjusts each sample’s contribution to the overall loss by evaluating the degree to which the rationale supports the answer. This mechanism directs the model to focus more on high-quality training examples that demonstrate more reliable reasoning processes and stronger alignment between the rationale and the answer. 

Experiments demonstrate that CoPeD outperforms the baselines on both IND and OOD benchmark datasets. Our contributions can be summarized as follows:

\begin{itemize}
    \item 	We propose a correctness-aware task setting where the model is trained to answer questions based on correct rationales and to revise erroneous  rationales. This design improves the consistency between the generated rationales and answers, thereby enhancing the faithfulness and soundness of reasoning.

    \item We develop a Correctness-Aware Confidence-Weighted Loss, which jointly considers rationale and answer prediction losses to re-weight training examples. This loss encourages the model to focus more on informative, well-aligned samples, while reducing the impact of noisy or misleading ones.
    \item We conduct comprehensive experiments across IND and OOD benchmarks, demonstrating that CoPeD effectively improves the reasoning performance of SLMs. 
    % The results show that our method improves the model’s generalization ability and reduces the negative impact of shortcut learning.
\end{itemize}

\begin{figure*}[tp]
  \includegraphics[width=0.9\linewidth]{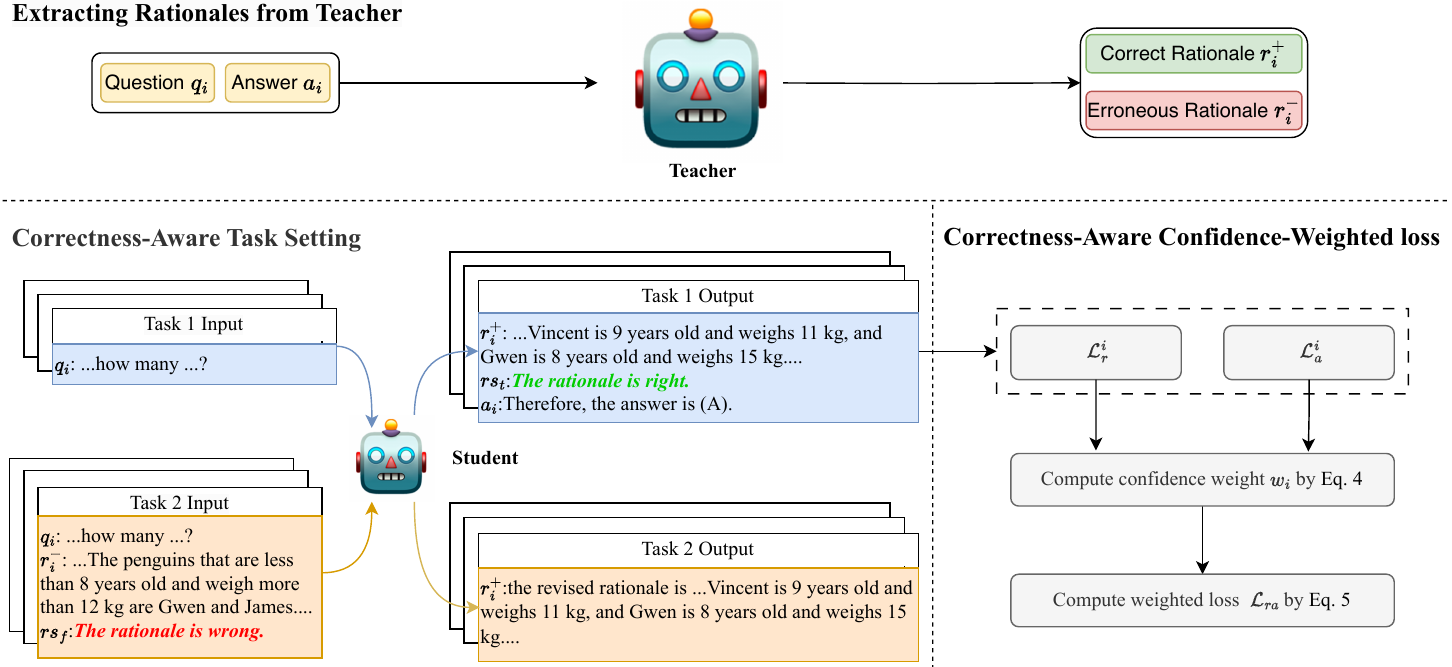}
      \vspace{-5pt}
	\centering
  \caption{Overview of \textbf{Chain-of-Thought Correctness Perception Distillation(CoPeD)}. We use teacher and student models to generate correct and erroneous rationales for the entire training set. Then, we adopt a multi-task learning framework to leverage these rationales, where one task is trained to predict the answer based on correct rationales, and the other task is trained to correct erroneous rationales as additional supervision signals.}
  \label{fig:CoPeD}
\end{figure*}

\section{Method}

The core idea of our method consists of two components: (1) guiding the student model with different training tasks based on the correctness of the rationale; and (2) encouraging the model to prioritize learning from high-quality and logically consistent rationales during training. From the perspectives of task design and data utilization, our approach jointly enhances the faithfulness and soundness of the rationales generated by the student model. The overall framework of our method is illustrated in Figure~\ref{fig:CoPeD}. In this section, we provide a detailed explanation of the method and discuss the motivation behind it.

% In this section, we introduce the proposed Chain-of-Thought Correctness Perception Distillation (CoPeD) method, which trains the student model using different strategies based on the correctness of the rationale, as illustrated in \ref{fig:CoPeD}.  

% (predicting the answer when the rationale is correct)
% (correcting the rationale when it is erroneous)
\subsection{Extracting Rationales from Teacher}
% \noindent \textbf{CoTs Annotated by Teacher LLMs:} 

For each training data sample $\mathcal{D}_{\text {train }}=\left\{\left({q}_i, {a}_i\right)\right\}_{i=1}^n$, we first employ a prompting method to automatically extract correct and erroneous rationales from the teacher model. Specifically, if the LLM’s predicted answer matches the ground truth, the corresponding rationale is considered likely correct; otherwise, it is assumed to be erroneous. We collect these rationales for two main purposes: (1) to enable the student model to learn from correct rationales; and (2) to enable the student model to learn how to correct erroneous rationales.
This method utilizes a few annotated examples to guide the teacher in generating rationales for new instances \cite{wei2022chain}. To maintain the quality of generated CoT, we following \citet{dai2024beyond} and use its provided prompt templates to guild the teacher generate correct and erroneous rationales with similar reasoning paths but different conclusions. Eventually, we construct the dataset $\mathcal{D}_{\mathrm{train}} = \{({q}_{i}, {r}_{i}^{+}, {r}_{i}^{-}, {a}_{i})\}_{i=1}^{n}$ for the student model, where ${q}_{i}$ is a question, ${a}_{i}$ is an answer, ${r}_{i}^{+}$ is the correct rationale, and ${r}_{i}^{-}$ is the erroneous rationale.

\begin{figure}[tp]
  \includegraphics[width=0.90\linewidth]{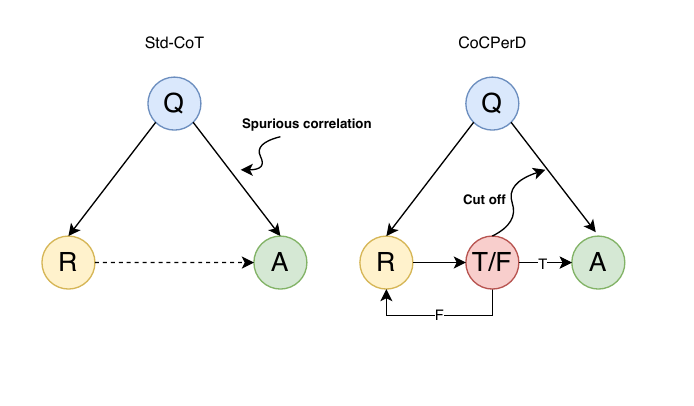}
    \vspace{-30pt}
	\centering
  \caption{
CoPeD adopts different strategies based on the correctness of the rationale, cutting off the spurious correlation between the question and the answer. 
% Here, $T$ represents the rationale status string $rs_t$, and $F$ represents $rs_f$.
}
  \label{fig:cut_off}
\end{figure}

\subsection{Correctness-Aware Task Setting}

To mitigate spurious correlations between questions and answers, we propose a correctness-aware task setting consisting of two tasks: answer prediction and rationale correction. To distinguish between these tasks, we append rationale status tokens ${rs}_{\mathrm{t}}$ (for correct rationale) and ${rs}_{\mathrm{f}}$ (for erroneous rationale) to the rationale. When the rationale is correct, the student model predicts the answer based on both the question and the rationale. When the rationale is erroneous, the model learns to revise the rationale. This framework encourages the model to rely on valid reasoning paths for answer prediction, rather than superficial question–answer correlations, thereby enhancing the faithfulness of the generated rationale. Additionally, the rationale correction task helps the model learn from mistakes. This reduces the probability of flawed reasoning steps during inference and improves the soundness of the generated rationale.

In the answer prediction task, the input to the student model is the question ${q}$, and its corresponding label consists of three components: the correct rationale  ${r}^+$, the rationale status string ${rs}_{\mathrm{t}}=\textbf{"the rationale is right"}$, and the answer ${a}$. The loss function for the answer prediction task is formulated as follows:
\begin{equation}
\small \mathcal{L}_{\text{ra}} = \mathbb{E}_{(q, r^+, a) \sim \mathcal{D}_{\mathrm{train}}} \left[ \ell(q, r^+ \oplus rs_{\mathrm{t}} \oplus a) \right]
\label{l_answer}
\end{equation}

In the rationale correction task, we concatenate the question ${q}$, the erroneous rationale ${r}^{-}$ and the rationale status string ${rs}_{\mathrm{f}}=\textbf{"the rationale is wrong"}$ as the input to the student model. The output label is the correct rationale $r^+$. This task design aims to enable the student model to learn to correct erroneous rationales. It thereby implicitly enhances the student model’s robustness and the quality of rationales generated during reasoning.
The loss function for the rationale correction task is formulated as follows:
\begin{equation}
\small
\mathcal{L}_{\text{rc}} = \mathbb{E}_{(q, r^+, r^-) \sim \mathcal{D}_{\mathrm{train}}} \left[ \ell(q \oplus r^- \oplus rs_{\mathrm{f}}, r^+) \right]
\label{l_rationale}
\end{equation}

The final objective function jointly optimizes the answer prediction loss $\mathcal{L}{\text{ra}}$ and the rationale correction loss $\mathcal{L}{\text{rc}}$, defined as:
\begin{equation}
\small \mathcal{L}_{\text {CoPeD}}=(1-\alpha) \mathcal{L}_{\text{ra }}+\alpha \mathcal{L}_{\text {rc}},
\end{equation}
where $\alpha$ is the hyperparameter used to weight the losses between the two learning tasks.

\subsection{Correctness-Aware Weighted loss}

While we initially assign rationale correctness labels based on whether the teacher model’s predicted answer matches the ground truth, this heuristic labeling may be noisy in practice.
Specifically, a rationale might contain logical flaws despite leading to a correct answer. Conversely, an erroneous answer might be supported by a seemingly plausible rationale.
Such coarse-grained supervision can mislead the student model during training. It may cause the model to overfit to unreliable or spurious reasoning paths.

To mitigate this issue, we propose a Correctness-Aware Weighted Loss, which dynamically adjusts the training contribution of each sample based on the degree of alignment between its rationale and answer. This mechanism enables the model to prioritize learning from samples with faithful and consistent rationale–answer pairs. At the same time, it down-weights samples exhibiting reasoning flaws or misalignment. By introducing this label-robust supervision strategy, the student model can better discern and rely on high-quality reasoning paths during distillation.

Concretely, we compute the rationale generation loss $\mathcal{L}_r$ and the answer prediction loss $\mathcal{L}_a$ for each instance. Based on these losses, we calculate a normalized confidence weight $w_i$, reflecting the reliability of the sample. Specifically, $w_i$ is computed as follows:
\begin{equation}
\small
    w_i = \text{softmax}_i ( 
        - \frac{ \mathcal{L}_r^{(i)} + \mathcal{L}_a^{(i)} + | \mathcal{L}_r^{(i)} - \mathcal{L}_a^{(i)} | }{ \tau } 
    ),
    \label{eq:wi}
\end{equation}
where composite loss term $\mathcal{L}_r + \mathcal{L}_a$ reflects the overall reliability of a sample. A high rationale loss indicates a noisy reasoning process. In contrast, a high answer prediction loss suggests that the rationale fails to support the final prediction. Furthermore, we use a discrepancy term $|\mathcal{L}_r - \mathcal{L}_a|$ to measure the alignment between the two objectives. The temperature parameter $\tau$ controls the smoothness of the resulting weight distribution.

The Correctness-Aware Weighted loss is then defined as a weighted summation over the sample-wise rationale and answer losses:
\begin{equation}
    \small \mathcal{L}_{ra} = \sum_i w_i \cdot ( \mathcal{L}_r^{(i)} + \mathcal{L}_a^{(i)} )
    \label{eq:weight_loss}
\end{equation}

To stabilize training, we initially use uniform weights during early epochs to allow the model to acquire basic reasoning capabilities. Starting from epoch $n$, the correctness-aware weighting mechanism is introduced to emphasize trustworthy samples adaptively. The complete training algorithm is presented in Algorithm~\ref{alg:correctness_weighting}.

\begin{algorithm}[t]
\caption{Training with Correctness-Aware Weighted loss}
\label{alg:correctness_weighting}
\small
\KwIn{Training dataset $\mathcal{D}$, student model $f_\theta$, temperature $\tau$, starting epoch $n$, cross entropy loss $CE(·, ·)$.}
\For{epoch $= 1$ to $N$}{
  \For{each mini-batch $\mathcal{B} = \{(q_i, a_i, r_i)\}_{i=1}^B$}{
    \For{each sample $(q_i, a_i, r_i) \in \mathcal{B}$}{
      Generate rationale $\hat{r}_i$ and predict answer $\hat{a}_i$: \;
      $\hat{r}_i, \hat{a}_i = f_\theta(q_i)$\;
      Compute rationale loss: $\mathcal{L}_r^{(i)} = CE(\hat{r}_i, r_i)$\;
      Compute answer loss: $\mathcal{L}_a^{(i)} = CE(\hat{a}_i, a_i)$\;
    }
    \uIf{epoch $< n$}{
      Compute unweighted loss: $\mathcal{L}_{ra} = \sum_i ( \mathcal{L}_r^{(i)} + \mathcal{L}_a^{(i)} )$\;
    }
    \Else{
      \For{each sample $i$ in $\mathcal{B}$}{
        Compute confidence weight $w_i$ by Eq.~\ref{eq:wi}\;
      }
      Compute weighted loss ${\mathcal{L}}_{ra}$ by Eq.~\ref{eq:weight_loss}\;
    }
    Update student model parameters using gradient descent;
  }
}
\end{algorithm}

% Intuitively, training samples with more faithful and consistent rationale-answer pairs should contribute more to model updates. In contrast, samples that exhibit either high rationale loss or poor answer prediction are considered less reliable and thus downweighted. This adaptive weighting encourages the model to prioritize learning from high-quality reasoning paths.

% Similarly, we compute a normalized confidence weight $w_i$ for each instance in the rationale correction task based on the correction loss $\mathcal{L}_{rc}$. The final loss is then calculated as a weighted sum:
% \begin{equation}
%     \small w_i = \frac{\exp\left(- \mathcal{L}_{rc}^{(i)} / T\right)}{\sum_j \exp\left(-\mathcal{L}_{rc}^{(j)}  / T\right)}
% \end{equation}
% \begin{equation}
%     \small \hat{\mathcal{L}}_{rc} = \sum_i w_i \cdot \mathcal{L}_{rc}^{(i)}
% \end{equation}

% Intuitively, instances with lower correction loss are assigned higher weights, encouraging the model to focus more on learnable and coherent rationale correction samples.

\begin{table*}[t]
\resizebox{\linewidth}{!}{
\begin{tabular}{lccccccc|c}
\toprule
\textbf{Method}    & \textbf{Distill?} & \textbf{Gen CoT?}  & \textbf{BBH-test} & \textbf{BB-sub} & \textbf{AGIEval} & \textbf{ARC-E} & \textbf{\hspace{0.5em}ARC-C\hspace{0.5em}} & \multirow{2}{*}{\textbf{\hspace{0.5em}AVG\hspace{0.5em}}} \\ \cmidrule{1-8}
In-domain?          & \multicolumn{1}{|c}{} & \multicolumn{1}{c|}{}  & \checkmark                 & \ding{53}                     & \ding{53}                & \ding{53}              & \ding{53}              &                               \\ \midrule
\multicolumn{9}{c}{\textbf{Teacher: ChatGPT (gpt-3.5-turbo)}}                                                                                                                                      \\ \midrule
Zero-shot-CoT & \multicolumn{1}{|c}{\ding{53}}& \multicolumn{1}{c|}{\checkmark} & 42.6              & 44.5                  & 50.3             & 92.1           & 82.2           & 62.3 \\ \midrule

 \multicolumn{9}{c}{\textbf{Student: LLaMA2-7B}}                                                                                                                                      \\ \midrule
% Zero-shot$^\spadesuit$     & \multicolumn{1}{|c}{\ding{53}}& \multicolumn{1}{c|}{\ding{53}} & 14.8              & 15.5                  & 6.9              & 18.2           & 13.9           & 13.9\\
% Zero-shot-CoT$^\spadesuit$ & \multicolumn{1}{|c}{\ding{53}}& \multicolumn{1}{c|}{\checkmark} & 10.6              & 7.7                   & 7.1              & 18.4           & 14.8           & 11.7\\  \midrule
% Few-shot$^\spadesuit$     & \multicolumn{1}{|c}{\ding{53}}& \multicolumn{1}{c|}{\ding{53}} & 15.1              & 28.5                  & 25.5             & 25.5           & 25.4           & 24.0\\
% Few-shot-CoT$^\spadesuit$ & \multicolumn{1}{|c}{\ding{53}}& \multicolumn{1}{c|}{\checkmark} & 16.3              & 25.3                  & 9.9              & 17.2           & 17.2           & 17.2\\
% Answer-SFT & \multicolumn{1}{|c}{\ding{53}}& \multicolumn{1}{c|}{\ding{53}} & 51.2             & 33.6                 & 30.8             & 72.1          & 53.5           & 48.2\\ \midrule
Std-CoT \cite{magister2023teaching}       & \multicolumn{1}{|c}{\checkmark}& \multicolumn{1}{c|}{\checkmark} & 58.5              & 29.5                  & 24.2             & 61.8           & 47.3           & 44.3\\
SCOTT \cite{wang2023scott}             & \multicolumn{1}{|c}{\checkmark}& \multicolumn{1}{c|}{\checkmark} & 43.1              & 19.7                  & 12.8& 46.3           & 35.9           & 31.6                          \\
MT-CoT \cite{li2022explanations}            & \multicolumn{1}{|c}{\checkmark} & \multicolumn{1}{c|}{\checkmark}& 59.3 & 31.4& 23.2& 51.7           & 40.6           & 41.2\\

Step-by-step \cite{hsieh2023distilling}      & \multicolumn{1}{|c}{\checkmark} & \multicolumn{1}{c|}{\checkmark}& 44.6              & 29.2                  & {28.4}    & 69.0& 49.2           & 43.2                          \\
CasCoD \cite{dai-etal-2024-improve-students}       & \multicolumn{1}{|c}{\checkmark}& \multicolumn{1}{c|}{\checkmark} & 60.2     & 37.2           & 28.6 & \underline{71.1} & \underline{52.4} & 49.9  \\     \midrule

CoPeD-T (ours)  & \multicolumn{1}{|c}{\checkmark}& \multicolumn{1}{c|}{\checkmark} & \underline{63.1}     & \underline{38.3}           & \underline{30.2} & \textbf{72.6} & \textbf{55.1}  & \underline{51.8}     \\
CoPeD-L (ours)  & \multicolumn{1}{|c}{\checkmark}& \multicolumn{1}{c|}{\checkmark} & 60.9     & 38.2           & 27.9 & 69.2 & 50.9 & 49.4     \\
CoPeD-TL (ours)  & \multicolumn{1}{|c}{\checkmark}& \multicolumn{1}{c|}{\checkmark} & \textbf{69.8}     & \textbf{39.5}           & \textbf{31.7} & \underline{71.1} & 52.3  & \textbf{52.9}  \\

\midrule
 \multicolumn{9}{c}{\textbf{Student: Mistral-7B-v0.2}}                                                                                                                                      \\ \midrule
Std-CoT\cite{magister2023teaching} & \multicolumn{1}{|c}{\checkmark}& \multicolumn{1}{c|}{\checkmark} & {72.5}& 36.8& {32.5}& 67.6& 58.6&53.6  \\

SCOTT \cite{wang2023scott}             & \multicolumn{1}{|c}{\checkmark}& \multicolumn{1}{c|}{\checkmark}   &31.9   &32.8   &27.3   &54.2   &38.6   &37.0   \\
MT-CoT \cite{li2022explanations}            & \multicolumn{1}{|c}{\checkmark} & \multicolumn{1}{c|}{\checkmark} &56.1   &39.4   &31.3   &68.4   &59.3   &50.9   \\

Step-by-step\cite{hsieh2023distilling}  & \multicolumn{1}{|c}{\checkmark}& \multicolumn{1}{c|}{\checkmark} & 58.1& {37.5}& 22.9& {78.4}& {61.7}&51.7 \\ % Added missing AVG value
CasCoD\cite{dai-etal-2024-improve-students}  & \multicolumn{1}{|c}{\checkmark}& \multicolumn{1}{c|}{\checkmark} & {70.5}& {39.5}& \underline{38.2}& \textbf{84.2}& \textbf{75.5}&\textbf{61.6} \\
\midrule    

CoPeD-T (ours)  & \multicolumn{1}{|c}{\checkmark}& \multicolumn{1}{c|}{\checkmark} & \underline{74.4}     & {40.7}           & {36.7} & 78.9 & 68.5  & 59.8  \\

CoPeD-L (ours)  & \multicolumn{1}{|c}{\checkmark}& \multicolumn{1}{c|}{\checkmark} & 74.1     & \underline{40.8}           & 36.1 & 80.1 & 65.4 &59.3      \\

CoPeD-TL (ours)  & \multicolumn{1}{|c}{\checkmark}& \multicolumn{1}{c|}{\checkmark} & \textbf{75.2}     & \textbf{41.2}           & \textbf{38.5} & \underline{82.6} & \underline{68.6}  & \underline{61.2}  

\\\bottomrule
\end{tabular}
}
\caption{Accuracy (\%) on in-domain and out-of-domain datasets with different methods. $^\spadesuit$: the results borrowed from \citet{dai-etal-2024-improve-students}. The best performance among distilled student models is marked in \textbf{bold}, and the  second-best performance is indicated by an \underline{underline}. CoPeD-T denotes the correctness-aware task setting, CoPeD-L refers to training Std-CoT using a correctness-aware weighted loss, and CoPeD-TL represents the combination of both methods.}
\label{tab:main-result}
\end{table*}

\section{Experiments}
In this section, we conduct extensive experiments and analyses to evaluate the effectiveness of our method on both in-domain (IND) and out-of-domain (OOD) datasets.

\subsection{Datasets}

\paragraph{In-domain Dataset:}
\textbf{BIG-Bench Hard (BBH)} \cite{suzgun2023challenging} consists of 27 challenging tasks drawn from BIG-Bench (BB) \cite{guobeyond}, covering domains such as arithmetic, symbolic reasoning, and others. Most tasks are multiple-choice questions, with a few open-ended ones. Following \citet{dai-etal-2024-improve-students}, we randomly split the BBH dataset into a training set (BBH-train) for distillation and a test set (BBH-test) for IND evaluation, using a 4:1 split.

\paragraph{Out-of-domain Dataset:} \textbf{(1) BIG-Bench Sub (BB-sub)} is derived from BIG-Bench (BB) \cite{guobeyond}, encompassing 203 tasks across domains such as linguistics, mathematics, and common-sense reasoning. To simplify our evaluation, we use the BB-Sub filtered by \citet{dai-etal-2024-improve-students}.
\textbf{(2) AGIEval} \cite{zhong2024agieval} is a benchmark that assesses language models (LMs) on reasoning abilities using human exams from fields including English, Mathematics, Law, and Logic. We select the English multiple-choice question subtask filtered by \citet{dai-etal-2024-improve-students}.
\textbf{(3) AI2 Reasoning Challenge (ARC)} \cite{Clark2018ThinkYH} consists of two datasets: ARC-Easy and ARC-Challenge, derived from middle and high school science exams. ARC-E features easier questions, while ARC-C presents more challenging ones. Following \citet{dai-etal-2024-improve-students}, we use the test sets from both datasets for evaluation.

\subsection{Implementation Details}
\paragraph{Models} We use LLaMA-7B \cite{Touvron2023Llama2O} as the base student model throughout all experiments unless otherwise specified. Given its cost-effectiveness and capabilities, we leverage OpenAI’s powerful black-box LLM, gpt-3.5-turbo-0613, as the teacher model to extract chain-of-thoughts (CoTs) using the same manual prompt as in prior works \cite{ dai2024beyond}.

\paragraph{Setup} We use LoRA \cite{hulora} for parameter-efficient fine-tuning of the student model. To balance the answer prediction and rationale correction tasks, we set $\alpha$ to 0.5. All experiments are performed using a mixed-precision training strategy on 8 × A800 GPUs. During inference, we utilize vLLM3 \cite{kwon2023efficient} to accelerate the process, employing a greedy decoding strategy for text generation on a single A800 GPU. Further details on training and hyperparameters are provided in Appendix \ref{appendix:hyperparameter}.

\paragraph{Baselines} We compare our method with the following baselines: (1) \textbf{Teacher} in  Zero-shot-CoT\cite{kojima2022large} for showing the impact of distilling reasoning ability from LLMs. (2) \textbf{Std-CoT} \cite{magister2023teaching}, which is the standard CoTs distillation method that directly fine-tune student models on the CoTs data.
%(3) \textbf{Step-by-step} \cite{hsieh2023distilling} is a multi-task CoTs distillation method that distills rationales and answers separately. 
(3) \textbf{MT-CoT} \cite{li2022explanations} is also a multi-task CoTs distillation method, but unlike Step-by-step, it simultaneously optimizes the objectives of answer prediction and entire CoTs learning. 
(4) \textbf{Step-by-step} \cite{hsieh2023distilling}  is a multi-task CoTs distillation method that distills rationales and answers separately
(5) \textbf{SCOTT} \cite{wang2023scott} that enhances the reasoning consistency of the student model by introducing additional counterfactual data.
% (5) \textbf{EDIT} \cite{dai2024beyond} uses prompts to generate dual CoTs data with similar reasoning paths but different conclusions, then applies the minimum edit distance algorithm to locate and optimize key reasoning steps.
% (7) \textbf{Std-CoT w/ Dual CoTs } \cite{dai2024beyond} train the Std-CoT using all data included in EDIT, adding the marker "[Counterfactual Reasoning]" before the negative sample’s question to differentiate it from positive reasoning.
(6) \textbf{CasCoD} \cite{dai-etal-2024-improve-students}  splitting single-step learning into two cascaded steps, restructuring training objectives to enhancing reasoning generalizability.

\subsection{Main Results}

% As shown in Table \ref{tab:main-result}, CoPeD-TL demonstrates competitive performance against strong baselines on both in-domain (IND) and out-of-domain (OOD) benchmarks. Specifically, LLaMA2-7B equipped with CoPeD-TL achieves an average accuracy of 52.9\% across all tasks, outperforming the strongest baseline, CasCoD, by 3.0\%. Notably, CoPeD-TL exhibits strong genealization in OOD scenarios, surpassing CasCoD by 2.3\%, 3.1\%, 1.5\%, and 2.7\% on BB-test, AGIEval, ARC-E, and ARC-C, respectively.

As shown in Table \ref{tab:main-result}, CoPeD demonstrates competitive performance against strong baselines on both in-domain (IND) and out-of-domain (OOD) benchmarks. Specifically, LLaMA2-7B equipped with CoPeD-TL achieves an average accuracy of 52.9\% across all tasks. It outperforms the  strongest baseline, CasCoD, by 3.0\%. In particular, it surpasses CasCoD by 9.6\% in the IND scenario. Meanwhile, CoPeD-TL also exhibits strong generalization ability in OOD scenarios. It outperforms CasCoD by 2.3\%, 3.1\%, 1.5\%, and 2.7\% on BB-test, AGIEval, ARC-E, and ARC-C, respectively.
While Mistral-7B with CoPeD-TL does not achieve the best results on every individual OOD scenario, it still delivers consistently competitive performance across a wide range of tasks.  This highlights the robustness and generalization ability of our method, even in comparison with state-of-the-art models.

These results validate the effectiveness of our design. CoPeD-T enables the student model to discern faithful reasoning paths from spurious ones, thereby improving both the quality and reliability of reasoning. Meanwhile, compared to the Std-CoT approach, CoPeD-L enhances training robustness by adaptively down-weighting misaligned or unreliable samples. It emphasizes high-quality, well-aligned rationales, allowing the model to focus on trustworthy reasoning paths.
% Together, these components demonstrate that CoPeD-TL enables the student model to better leverage correct rationales during learning, leading to faithful reasoning and robust performance across diverse tasks and domains.

\subsection{Faithfulness and Soundness of Students}

Inspired by previous work \cite{wang2023scott, dai-etal-2024-improve-students}, we employ LLMs as evaluators to assess two aspects. First, whether the rationale provided by the student model supports its prediction (i.e., faithfulness). Second, whether the rationale supports the ground-truth answer (i.e., soundness). Given a rationale $\hat{r}_{i}$ generated by the student model and an answer (either the predicted answer $\hat{a}_{i}$ or the ground-truth answer ${a}_i$), we construct evaluation prompt\footnote{The prompt for evaluating whether the rationale provided by the student model supports the answer can be found in the Appendix \ref{appendix:Faithfulness CoTs}} $p_{e}$  to guide LLM-based scoring. We define faithfulness and soundness as follows:
\begin{align}
\small \text{{Faithfulness}} &= \mathbb{E} \big[ f_{\mathrm{eval}}(p_e,q_i,\hat{r}_i, \hat{a}_i) \big], \\
\small \text{{Soundness}} &= \mathbb{E} \big[ f_{\mathrm{eval}}(p_e,q_i,\hat{r}_i, {a}_i) \big],
\end{align}
where  $f_{\mathrm{eval}}(\hat{r}_i, \hat{a}_i)$ and  $f_{\mathrm{eval}}(\hat{r}_i, {a}_i) \in \{0,1\}$ are a binary evaluation function, returning 1 if the rationale $\hat{r}_i$ sufficiently supports the given answer (either the predicted answer $\hat{a}_i$ or the ground-truth answer ${a}_i$), and 0 otherwise.

\begin{table}[htb]
\resizebox{\linewidth}{!}{
\begin{tabular}{c|ccc|ccc}
\hline
\multirow{2}{*}{\textbf{Method}} & \multicolumn{3}{c|}{\textbf{Faithfulness}}                     & \multicolumn{3}{c}{\textbf{Soundness}}                      \\ \cline{2-7}
                        & \textbf{ChatGPT} & \multicolumn{1}{c|}{\textbf{GPT4}} & \textbf{AVG}  & \textbf{ChatGPT} & \multicolumn{1}{c|}{GPT4} & \textbf{AVG}  \\ \hline
Teacher                 & 86.6        &  \multicolumn{1}{c|}{86.9}     & 86.8     & 74.8         &  \multicolumn{1}{c|}{71.5}     & 73.2     \\ \hline
Std-CoT                 & 80.5    & \multicolumn{1}{c|}{67.9} & 74.2 & 64.0    & \multicolumn{1}{c|}{54.5} & 59.3 \\
CasCoD                  & 82.2    & \multicolumn{1}{c|}{72.6} & 77.4 & 70.2    & \multicolumn{1}{c|}{59.6} & 64.9 \\
CoPeD-TL           & 83.8    & \multicolumn{1}{c|}{78.5} & 81.2  & 72.6    & \multicolumn{1}{c|}{67.9} &70.2  \\ \hline
\end{tabular}
}
\caption{Faithfulness (\%) and Soundness (\%) of the compared methods on the IND dataset. We employ both ChatGPT and GPT-4 as evaluators to mitigate the risk of single-model bias.}
\label{tab:faithfulness}
\end{table}

The results are shown in Table \ref{tab:faithfulness}. Compared to the baseline, the rationale generated by CoPeD-TL is more consistent with the answer. This includes both the predicted and the ground-truth answers.
This indicates that CoPeD-TL ensures the faithfulness and soundness of the rationale generated during the reasoning process. It does so by adopting different strategies based on the correctness of the rationale and filtering noisy samples. This approach helps mitigate spurious correlations between the question and the answer.

\begin{figure*}[hbt]
	\centering
	\includegraphics[width=\linewidth]{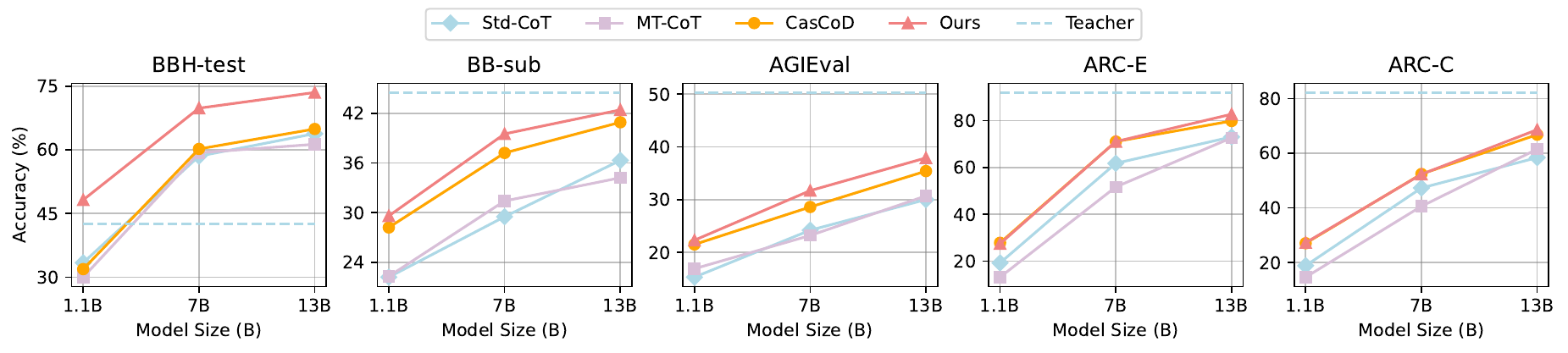}
    \vspace{-20pt}
	\centering
	\caption{Ablation study on model size for IND and  four OOD datasets. The dotted line indicates the performance of the teacher LLM under the Zero-shot-CoT setting.}
	\label{ablation-on-model-size-ood}
\end{figure*}

\begin{figure*}[hbt]
	\centering
	\includegraphics[width=\linewidth]{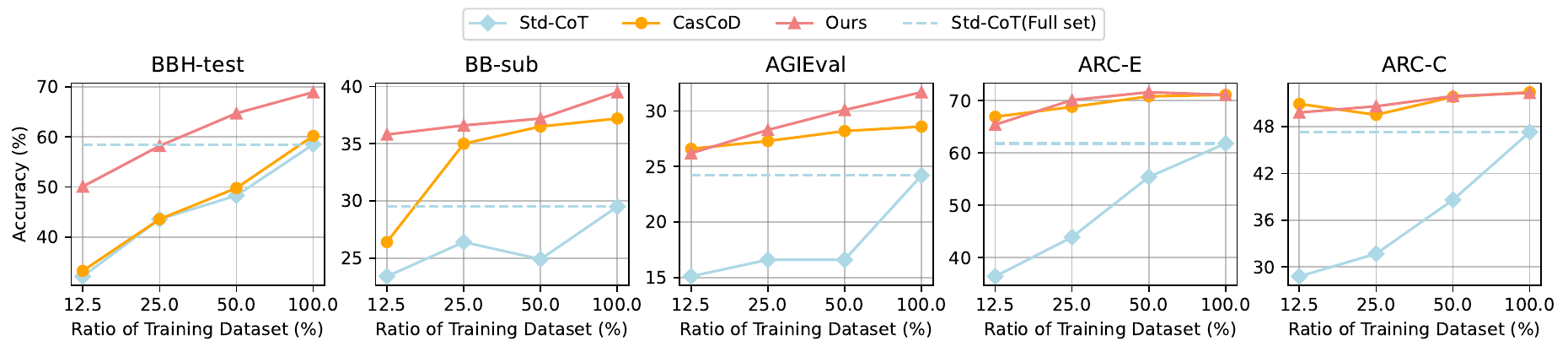}
    \vspace{-20pt}
	\centering
	\caption{Ablation study on training data size for IND and four OOD datasets. The dotted line indicates the performance of fine-tuning the student models by Std-CoTs distillation using the full set (100\% of) BBH-train dataset.}
	\label{ablation-on-train-data-size-ood}
\end{figure*}

\subsection{Ablation Study}

\paragraph{Model Size}
We conducted model distillation on TinyLLaMA-1.1B\footnote{\url{https://huggingface.co/TinyLlama/TinyLlama-1.1B-intermediate-step-1431k-3T}} \cite{zhang2024tinyllama}, LLaMA2-7B, and LLaMA2-13B, and compared it with Std-CoT, MT-CoT, and CasCod. As shown in Figure \ref{ablation-on-model-size-ood}, CoPeD-TL consistently achieves competitive performance across student models of varying sizes, outperforming baseline methods on both IND and OOD datasets. Notably, on the IND dataset, the 1.1B model with CoPeD-TL reaches 113.1\% of the teacher model’s performance, demonstrating the significant advantages of CoPeD-TL in low-resource scenarios. Moreover, across different model scales, CoPeD-TL maintains competitive performance on OOD datasets compared to baseline approaches.

\paragraph{Data Size}
CoPeD-TL demonstrates significant improvements over baseline methods on both IND and OOD datasets, while utilizing considerably less training data. As shown in Figure \ref{ablation-on-train-data-size-ood}, CoPeD-TL achieves a 16.8\% improvement over CasCoD on the IND (BBH-test) dataset, using only 12.5\% of the full BBH-train data. The performance on OOD datasets is even more notable. For example, on the BB-sub dataset, CoPeD-TL achieves a 9.4\% improvement in accuracy compared to CasCoD, even when using only 12.5\% of the complete BBH-train data.
On other OOD datasets, CoPeD-TL also achieves excellent performance. These results clearly demonstrate the effectiveness of CoPeD-TL in low-resource settings. They highlight its ability to enhance the performance of CoTs in both IND and OOD scenarios while requiring significantly less training data.

\begin{table}[t]
\centering
\small
\resizebox{\linewidth}{!}{
\begin{tabular}{lcc}
\toprule
\textbf{Method} & \textbf{Accuracy (\%)} & \textbf{Gain  (\%)} \\
\midrule
Uniform Weight (Std-CoT) & 58.5 & – \\
Only Composite Term ($\mathcal{L}_r + \mathcal{L}_a$) & 60.2 & +1.7 \\
Only Discrepancy Term ($|\mathcal{L}_r - \mathcal{L}_a|$) & 59.8 & +1.3 \\
\textbf{Full (CoPeD-L)} & \textbf{60.9} & \textbf{+2.4} \\
\bottomrule
\end{tabular}
}
\caption{Ablation results of different loss weighting strategies on BBH-test. }

\label{tab:ablation-loss-term}
\end{table}

\paragraph{Loss Term}
To evaluate the individual contributions of the components within our correctness-aware weighted loss, we conduct an ablation study focusing on the BBH-test accuracy, as shown in Table~\ref{tab:ablation-loss-term}. When only the composite loss term $(\mathcal{L}_r + \mathcal{L}_a)$ is used to compute sample weights, the model achieves a moderate improvement over the uniform weighting baseline. This suggests that considering the overall sample difficulty helps filter out noisy examples. Similarly, utilizing only the discrepancy term $|\mathcal{L}_r - \mathcal{L}_a|$ yields a modest accuracy gain. This indicates that rationale–answer alignment alone also provides useful supervision. Importantly, combining both components in the full weighting scheme leads to a substantial boost in performance. These results highlight the importance of modeling both sample quality and rationale–answer consistency, enabling the student model to focus on more trustworthy reasoning trajectories.

\subsection{Analysis \& Case Study}
% 由于页数限制，我们在附录提供了系统地案例研究来阐述CoT泛化性的提升.
% Due to page limitations, we provide a systematic case study in Appendix \ref{appendix:case-study} to illustrate the improvement in CoT generalizability.

Due to page limitations, we provide analysis in Appendix \ref{outline-analysis} and case study in Appendix \ref{appendix:case-study}.

\section{Related Works}

\paragraph{Chain-of-Thought Distillation}

Recent studies have demonstrated that CoT prompts significantly enhance the reasoning ability of LLMs for complex tasks \cite{wei2022chain, kojima2022large, wangself, huang2023large}. However, this advantage is most pronounced in LLMs, prompting several researchers \cite{magister2023teaching, ho2023large, li2023symbolic, chae2023dialogue, yang2024effective} to explore methods for transferring reasoning knowledge from LLMs to SLMs. Typically, these approaches leverage CoT prompts to generate rationales from LLMs, which are then used to fine-tune SLMs.

In addition, \citet{hsieh2023distilling} argue that reasoning bases and answers should be treated as distinct optimization objectives. Similarly, \citet{li2022explanations} suggest that learning both the complete CoT and individual answers can enhance the reasoning capabilities of the student model. \citet{liu2024mind} introduce an additional distillation objective focused on self-assessment, enabling the SLM to evaluate the accuracy of its generated CoTs. \citet{wang2023scott} propose reducing reasoning errors and hallucinations inherited by the SLM from the LLM through contrastive decoding, which ensures that the reasoning basis is closely related to the answer. Moreover, \citet{wang2023democratizing} present an interactive, multi-turn paradigm that allows the SLM to engage in self-reflection and receive feedback from the LLM during the learning process. \citet{dai-etal-2024-improve-students} suggest decomposing the traditional single-step learning process into two cascading steps to alleviate the effects of spurious correlations between questions and answers. \citet{lee2024mentor} effectively enhances the reasoning ability of small models by introducing an intermediate-sized, task-specific “mentor” model to improve the quality of multi-step reasoning distillation and provide soft labels. \citet{feng2024teaching} proposes a counterfactual distillation framework that improves the reasoning ability and OOD robustness of small language models. \citet{wadhwa2024investigating} investigates why CoT rationales help in model distillation and finds that even incoherent or partial rationales appended after labels can significantly improve student model performance. \citet{chenglin2024mixed} proposed a Mixed Distillation framework that efficiently distills multi-step reasoning abilities into small models through multi-path reasoning samples and multi-task loss.
\citet{zhu2024pad} proposes Program-aided Distillation (PaD), which improves the distillation quality of reasoning tasks by using reasoning programs to correct errors in synthetic data and iteratively refine the distilled model’s reasoning capabilities.

% In contrast, our method enables the student model to recognize that answer prediction should be based on a valid reasoning path, thereby mitigating the spurious correlation between the question and the answer.

\paragraph{Learning from Mistakes}
Recent studies have investigated the use of mistake data to improve the performance of language models. \citet{shinn2024reflexion} introduce Reflexion, a method that allows LLM agents to self-reflect on their mistakes. \citet{wang2023learning} propose a study assistant that collects and retrieves training mistakes from LLMs to guide future inferences. \citet{lichain} present the CoK method, which corrects reasoning errors by retrieving relevant knowledge to prevent the propagation of errors. However, these approaches are not directly applicable to standard SLMs. \citet{wang2023scott} propose fine-tuning on counterfactual data to ensure the faithful reasoning of the student model. \citet{an2023learning} introduce LEMA, a method that fine-tunes language models on corrected mistake data, with mistakes collected from various LLMs. 
\citet{tong2024can} explores whether large language models (LLMs) can enhance their reasoning abilities by learning from their mistakes, proposing two methods—self-rethinking prompting and mistake tuning. 
\citet{an2024can} investigates whether large language models (LLMs) can enhance their CoT reasoning by learning from mistakes.
% However, these methods overlook the fact that the task of correcting erroneous reasoning can implicitly improve the reasoning quality of SLMs.

\section{Conclusion}
In this study, we propose a Chain-of-Thought Correctness Perception Distillation framework (CoPeD).
It employs a dual-task training mechanism comprising answer prediction and rationale correction to significantly enhance the faithfulness and soundness of reasoning. To address the noise present in teacher-generated data, we introduce a correctness-aware weighted loss.
This loss effectively reduces the negative impact of unreliable samples and strengthens the model’s ability to identify and leverage high-quality reasoning paths. Extensive experiments across varying model sizes and training data volumes demonstrate that CoPeD consistently achieves superior performance on both IND and OOD benchmarks, validating its effectiveness in improving reasoning quality and generalization capability.
\newpage

\section{Limitations}

% While CoPeD demonstrates strong improvements in reasoning faithfulness and soundness, it has several limitations. First, our approach relies on correctness labels that are heuristically derived based on whether the predicted answer matches the ground truth. Although our weighted loss mitigates the impact of this coarse supervision, it may still introduce noise in certain edge cases. Second, the current weighting mechanism is relatively simple and may be further refined to better capture subtle inconsistencies between rationales and answers. 

In our study, we explore enabling the student model to verify the correctness of the generated rationale during inference and to attempt corrections when the rationale is identified as erroneous. However, the student model currently struggles to effectively validate whether the rationale derived from its reasoning is indeed correct. 
% Even assuming the model can accurately verify the rationale’s correctness during inference, its ability to recover from errors remains limited. This limitation arises because verifying and correcting rationale correctness constitutes a more complex reasoning task, which poses particular challenges for small language models (SLMs).

Moreover, our approach depends on correctness labels heuristically derived from whether the predicted answer matches the ground truth. Although our weighted loss function alleviates the impact of this coarse supervision, it may still introduce noise in certain edge cases. Additionally, the current weighting mechanism is relatively simple and could be further refined to better capture subtle inconsistencies between rationales and answers.

% \section*{Acknowledgments}

% This document has been adapted
% by Steven Bethard, Ryan Cotterell and Rui Yan
% from the instructions for earlier ACL and NAACL proceedings, including those for
% ACL 2019 by Douwe Kiela and Ivan Vuli\'{c},
% NAACL 2019 by Stephanie Lukin and Alla Roskovskaya,
% ACL 2018 by Shay Cohen, Kevin Gimpel, and Wei Lu,
% NAACL 2018 by Margaret Mitchell and Stephanie Lukin,
% Bib\TeX{} suggestions for (NA)ACL 2017/2018 from Jason Eisner,
% ACL 2017 by Dan Gildea and Min-Yen Kan,
% NAACL 2017 by Margaret Mitchell,
% ACL 2012 by Maggie Li and Michael White,
% ACL 2010 by Jing-Shin Chang and Philipp Koehn,
% ACL 2008 by Johanna D. Moore, Simone Teufel, James Allan, and Sadaoki Furui,
% ACL 2005 by Hwee Tou Ng and Kemal Oflazer,
% ACL 2002 by Eugene Charniak and Dekang Lin,
% and earlier ACL and EACL formats written by several people, including
% John Chen, Henry S. Thompson and Donald Walker.
% Additional elements were taken from the formatting instructions of the \emph{International Joint Conference on Artificial Intelligence} and the \emph{Conference on Computer Vision and Pattern Recognition}.

\bibliography{acl_latex}

\newpage

\appendix

\section{Additional Experiment Results}

\subsection{Ablation on Temperature $\tau$}

\begin{table}[ht]
    \centering
    \resizebox{\linewidth}{!}{
    \begin{tabular}{c|ccccc}
        \toprule
        Temperature $\tau$ & 1 & 2.5 & 5.0 & 7.5 & 10.0 \\
        \midrule
        BBH-test Accuracy (\%) & 64.8 & 69.2 & \textbf{69.8} & 68.5 & 66.1 \\
        \bottomrule
    \end{tabular}
    }
    \caption{Effect of temperature $\tau$ in the correctness-aware weighting loss on BBH-test accuracy.  }
    \label{tab:temperature}
\end{table}

We conduct an ablation study to investigate the impact of the temperature coefficient $\tau$ in our proposed CoPeD-TL. As shown in Table~\ref{tab:temperature}, we evaluate CoPeD-TLon the BBH-test set under various $\tau$ values.
The results show that performance increases as $\tau$ grows from 0.5 to 2.5, peaking at 69.8\%. Further increasing $\tau$ causes a slight decline, indicating a trade-off between weighting sharpness and generalization.
Specifically, small $\tau$ values lead to overconfident weighting on a few low-loss samples, which may introduce bias and hinder learning from diverse correct examples. Conversely, large $\tau$ makes the softmax distribution nearly flat, diminishing the benefit of correctness-aware weighting.

\begin{table*}[h]
\centering
\small
\begin{tabular}{lccccc|c}
\toprule
$\alpha$ & BBH-test & BB-sub & AGIEval & ARC-E & ARC-C & AVG \\
\midrule
0.3 & 70.4 & 37.7 & 28.1 & 69.7 & 51.4 & 51.5 \\
0.5 & 69.8 & 39.5 & 31.7 & 71.1 & 52.3 & \textbf{52.9} \\
0.7 & 63.5 & 38.6 & 29.7 & 70.2 & 50.7 &50.5 \\
\bottomrule
\end{tabular}
\caption{Ablation study on the task balance parameter $\alpha$. $\alpha$ controls the trade-off between answer prediction and rationale correction.}
\label{tab:alpha_ablation}
\end{table*}

\subsection{Ablation on Task Balance Parameter $\alpha$}

To investigate the impact of task balancing between rationale correction and answer prediction, we conduct an ablation study on the weighting parameter $\alpha$ used in our multi-objective loss. Specifically, we vary $\alpha \in {0.3, 0.5, 0.7}$ to control the relative importance of rationale supervision versus answer supervision. The results are reported in Table~\ref{tab:alpha_ablation}.

From the results, we observe that $\alpha = 0.5$ yields the best average performance across all benchmark datasets, indicating that a balanced emphasis on both rationale correction and answer prediction leads to more effective reasoning. When $\alpha$ is set to 0.3, the model prioritizes answer prediction while underutilizing the benefits of rationale supervision, resulting in suboptimal generalization. Conversely, when $\alpha$ is increased to 0.7, the model places excessive focus on rationale correction, potentially at the cost of answer prediction quality. These findings suggest that maintaining a moderate balance between the two tasks is crucial for achieving strong overall reasoning ability and robustness across diverse benchmarks. Therefore, we adopt $\alpha = 0.5$ as the default setting in our final model.

\section{Experimantal Settings}

\subsection{Hyperparameters Settings}
\label{appendix:hyperparameter}
To guarantee the fairness of our comparative analysis, in our study, we keep the hyperparameter settings consistent across all baselines, our proposed CoPeD approach included. Below, we provide a detailed account of the hyperparameter configurations used in our experiments.
The detailed hyperparameters in training and inference can be found in Table \ref{tab:train-hyperparameters} and Table \ref{tab:infer-hyperparameters}, respectively.

In our research, We maintain a consistent batch size across all baselines to eliminate performance differences caused by varying batch sizes. Through a series of experiments with learning rates set to 5e-5, 1e-4, 2e-4, 3e-4 and 4e-4, 5e-4 we find that the learning rate is a critical factor affecting model performance and that the optimal value varies with model size. Therefore, we adjust the learning rate accordingly based on model size.

\begin{table*}[!htb]
\scriptsize
\centering
\resizebox{\linewidth}{!}{
\begin{tabular}{l|cccc}
\toprule
\textbf{Hyperparameter} &  \textbf{TinyLLaMA-1.1B}&\textbf{LLaMA2-7B}& \textbf{LLaMA2-13B}  &\textbf{Mistral-7B-v0.2}\\ \midrule
gradient accumulation steps &   2   &   2   &   2   &   2\\
per device batch size   &  2    &   2   &   2   &   2\\
learning rate           &  4e-4 &   3e-4  &   1e-4  &   3e-4\\
epoches                 &  30   &   20  &    20 &    20\\
temperature $\tau$      &  2.5  &   5   &   5   &   5\\
starting epoch $n$      &  10   &   5   &   5   &   5\\

max length              &  1024             &1024             & 1024    & 1024              \\
$\beta$ of AdamW    &  (0.9,0.999)  &   (0.9,0.999) & (0.9,0.999)   & (0.9,0.999)\\
 $\epsilon$ of AdamW    &   1e-8  &   1e-8  &   1e-8    &   1e-8\\
$\gamma$ of Scheduler   &   0.95    &   0.95    &   0.95    &   0.95\\
weight decay            &  0                &0                & 0   & 0                 \\
warmup ratio            &  0                &0                & 0   & 0                \\ 
rank of LoRA    &   64  &   64  &   64  &   64\\ 
$\alpha$ of LoRA    &  32   &   32 & 32 & 32\\
 target modules & q\_proj, v\_proj  & q\_proj, v\_proj  &   q\_proj, v\_proj    &   q\_proj, v\_proj\\ 
drop out of LoRA    &  0.05 &0.05   & 0.05  & 0.05\\ 
\bottomrule
\end{tabular}
}
\caption{Training hyperparameters.}
\label{tab:train-hyperparameters}
\end{table*}

\begin{table}[!htb]
\footnotesize
\centering
\begin{tabular}{l|cc}
\toprule
\textbf{Arguments}&  \textbf{Student}&\textbf{Teacher}\\ \midrule
do sample&  False&True\\
temperature&  -&0.2\\
 top-p& 1.0&1.0\\
top-k&  -&-\\
max new tokens&  1024             &2048\\
\# return sequences&  1&1\\
\bottomrule
\end{tabular}
\caption{Generation configs of students and teachers.}
\label{tab:infer-hyperparameters}
\end{table}

\subsection{Dataset Statistics}
\label{appendix:data-stat}
Table \ref{tab:agieval_sat}, Table \ref{tab:arc_sat}, Table \ref{tab:bbheval_sat} and Table \ref{tab:bbsubeval_sat} show the data statistics of AGIEval, ARC, BIG-Bench Hard (BBH) and BIG-Bench Sub (BB-sub), respectively.
\begin{table}[!h]
\footnotesize
\centering
\begin{tabular}{l|l|c|c}
\toprule
 \textbf{No.}&\textbf{Task}                         & \textbf{Size}& \textbf{\# Choices} \\ \midrule
1&AQuA-RAT               & 254                  & 5                   \\
2&LogiQA-EN& 651                  & 4                   \\
3&LSAT-AR                & 230                  & 5                   \\
4&LSAT-LR                & 510                  & 5                   \\
5&LSAT-RC                & 269                  & 5                   \\
6&SAT-Math               & 220                  & 4                   \\
7&SAT-EN& 206                  & 4                   \\
8&SAT-EN (w/o Psg.)& 206                  & 4      \\     
\midrule
 &\textbf{Sum} & 2546& - \\\bottomrule         
\end{tabular}
\caption{Statistics of AGIEval dataset.}
\label{tab:agieval_sat}
\end{table}

\begin{table}[htbp]
\footnotesize
\centering
\begin{tabular}{l|c|c}
\toprule
\textbf{Task}          & \textbf{Size}& \textbf{\# Choices} \\ \midrule
ARC-E               & 2376                  & 4-5                   \\
ARC-C & 1172                  & 4-5                   \\   \bottomrule
\end{tabular}
\caption{Statistics of ARC test dataset.}
\label{tab:arc_sat}
\end{table}

\begin{table}[!htb]
\scriptsize
\centering
\begin{tabular}{l|l|c|c}
\toprule

\textbf{No.}&\textbf{Task}                         & \textbf{Size}& \textbf{\# Choices} \\ \midrule
1&Reasoning about Colored Objects       & 250                  & 18                  \\
2&Geometric Shapes                      & 250                  & 11                  \\
3&Ruin Names                            & 250                  & 11                   \\
4&Penguins in a Table                   & 146                  & 5                   \\
5&Movie Recommendation                  & 250                  & 5                   \\
6&Tracking Shuffled Objects (3 objects)& 250                  & 3\\
7&Tracking Shuffled Objects (5 objects)& 250                  & 5\\
8&Tracking Shuffled Objects (7 objects)& 250                  & 7\\
9&Logical Deduction (3 objects)& 250                  & 3\\
10&Logical Deduction (5 objects)& 250                  & 5\\
11&Logical Deduction (7 objects)& 250                  & 7\\
12&Date Understanding                    & 250                  & 6                   \\
13&Salient Translation Error Detection   & 250                  & 6                   \\
14&Causal Judgement                      & 187                  & 2                   \\
15&Disambiguation QA                     & 250                  & 4                   \\
16&Temporal Sequences                    & 250                  & 4                   \\
17&Boolean Expressions                   & 250                  & 2                   \\
18&Hyperbaton (Adjective Ordering)& 250                  & 2                   \\
19&Navigate                              & 250                  & 2                   \\
20&Snarks                                & 178                  & 2                   \\
21&Sports Understanding                  & 250                  & 2                   \\
22&Formal Fallacies Syllogisms Negation& 250                  & 2                   \\
23&Web of Lies                           & 250                  & 2 \\
24&Dyck Languages& 250&-\\
25&Multi-Step Arithmetic& 250&-\\
26&Object Counting& 250&-\\
27&Word Sorting& 250&-\\
\midrule
&\textbf{Sum} & 6511& - \\\bottomrule        
\end{tabular}
\caption{Statistics of BIG-Bench Hard dataset.}
\label{tab:bbheval_sat}
\end{table}

\begin{table}[!htb]
    \scriptsize
    \centering
    \begin{tabular}{l|l|c|c}
    \toprule
     \textbf{No.}&\textbf{Task}                         & \textbf{Size}& \textbf{\# Choices} \\ \midrule
     1 &abstract\_narrative\_understanding& 100& 5\\
     2 &anachronisms& 100& 2                   \\
     3 &analogical\_similarity& 100& 7\\
     4 &analytic\_entailment& 70& 2\\
     5 & cause\_and\_effect& 100&2\\
      6 &checkmate\_in\_one& 100&26\\
     7 &cifar10\_classification& 100& 10\\
     8 &code\_line\_description& 60& 4\\
     9 &conceptual\_combinations& 100& 4\\
     10 &crass\_ai& 44& 4\\
     11 &elementary\_math\_qa& 100& 5                   \\
     12 &emoji\_movie& 100& 5\\
      13 &empirical\_judgments& 99&3\\
     14&english\_russian\_proverbs& 80& 4\\
     15&entailed\_polarity& 100& 2\\
     16&entailed\_polarity\_hindi& 100& 2\\
     17&epistemic\_reasoning& 100& 2\\
     18&evaluating\_information\_essentiality& 68& 5\\
     19&fantasy\_reasoning& 100& 2                   \\
     20&figure\_of\_speech\_detection& 59& 10\\
     21
    &goal\_step\_wikihow
    & 100& 4\\ 
     22
    &gre\_reading\_comprehension
    & 31& 5\\
     23
    &human\_organs\_senses
    & 42& 4\\ 
     24
    &identify\_math\_theorems
    & 53& 4\\
     25
    &identify\_odd\_metaphor
    & 47& 5\\
     26& implicatures& 100&2\\ 
     27&implicit\_relations
    & 82& 25\\
     28&indic\_cause\_and\_effect
    & 100& 2 \\ 
     29&intersect\_geometry
    & 100& 26\\
     30&kanji\_ascii
    & 100& 5\\ 
     31&kannada
    & 100& 4\\
     32&key\_value\_maps
    & 100& 2 \\ 
     33&logic\_grid\_puzzle
    & 100& 3\\
     34&logical\_args
    & 32& 5\\ 
     35&logical\_fallacy\_detection
    & 100& 2\\
     36&metaphor\_boolean
    & 100& 2\\ 
     37&metaphor\_understanding
    & 100& 4\\ 
     38&minute\_mysteries\_qa
    & 100& 4\\
     39&mnist\_ascii
    & 100& 10\\ 
     40&moral\_permissibility
    & 100& 2\\
     41&movie\_dialog\_same\_or\_different
    & 100& 2 \\ 
     42&nonsense\_words\_grammar
    & 50& 4\\
     43&odd\_one\_out
    & 86& 5\\ 
     44&parsinlu\_qa
    & 100& 4\\
     45&physical\_intuition
    & 81& 4\\ 
     46&play\_dialog\_same\_or\_different
    & 100& 2\\
     47&presuppositions\_as\_nli
    & 100& 3\\ 
     48&riddle\_sense
    & 49& 5\\ 
     49&similarities\_abstraction
    & 76& 4\\
     50&simple\_ethical\_questions
    & 100& 4\\ 
     51&social\_iqa
    & 100& 3\\
     52&strange\_stories
    & 100& 2 \\ 
     53&strategyqa
    & 100& 2\\
     54&swahili\_english\_proverbs
    & 100& 4\\ 
     55&swedish\_to\_german\_proverbs
    & 72& 4\\
     56&symbol\_interpretation
    & 100& 5\\ 
     57&timedial
    & 100& 3\\
     58&undo\_permutation
    & 100& 5\\ 
     59&unit\_interpretation
    & 100& 5\\ 
     60&vitaminc\_fact\_verification
    & 100& 3\\
     61&winowhy& 100& 2 \\
    \midrule
     &\textbf{Sum} & 5384& - \\
     \bottomrule             
    \end{tabular}
    \caption{Statistics of BIG-Bench sub dataset. We filter the original dataset by retrieving tasks with keywords "multiple choice" and randomly sample up to 100 examples per task. Note, the task in BBH will not be involved in BB-sub.}
    \label{tab:bbsubeval_sat}
\end{table}

\section{Prompts}
\subsection{Prompts of Correct the Erroneous Rationale for ChatGPT}
\label{appendix:correct CoTs}
We use the prompt template shown in Table \ref{tab:prompt-gen-CoTs} to call the ChatGPT API to correct the erroneous rationale of student model for the BBH-train datasets.
\begin{table*}[h]
\small
\centering
\begin{tabular}{l|l}
\toprule
system content & \parbox[c]{13cm}{%
\texttt{You are a helpful and precise assistant for following the given instruction.}
}\\
\midrule 
user content & \parbox[c]{13cm}{%

\texttt{[Instruction]\{Please correct the wrong rationale by using better reasoning steps.\}}\\ \\
\texttt{Task Description:\{Task Description\}} \\ \\
\texttt{Question: \{Question\}}\\ \\
\texttt{Answer: \{Answer\}}\\ \\
\texttt{Wrong rationale: \{Wrong rationale\}}\\ \\
\texttt{Better Reasoning:}\\
}\\
\bottomrule
\end{tabular}
\caption{Prompt template for gpt-3.5-turbo for ask the teacher LLM to generate correct rationales.}
\label{tab:prompt-gen-CoTs}
\end{table*}

% \begin{table*}[h]
% \small
% \centering
% \begin{tabular}{l|l}
% \toprule
% system content & \parbox[c]{13cm}{%
% \texttt{You are a helpful and precise assistant for following the given instruction.}
% }\\
% \midrule 
% user content & \parbox[c]{13cm}{%
% \texttt{[Instruction] \{Please read the question and the rationale, and then give your answer based on the question and the rationale without any explanations.\}} \\ \\
% \texttt{Task Description: \{TASK\_DESCRIPTION\}}\\ \\
% \texttt{Question: \{QUESTION\}} \\ \\
% \texttt{Rationale: \{RATIONALE\}} \\ \\
% \texttt{Your Answer: } \\
% }\\
% \bottomrule
% \end{tabular}
% \caption{Prompt template of simulators for predicting the answers when given the question and rationale.}
% \label{tab:prompt-q-r-a}
% \end{table*}

\subsection{Prompts of Evaluator}
\label{appendix:Faithfulness CoTs}
We use the prompt templates shown in Table \ref{tab:evaluator-gen-CoTs} to call the ChatGPT and GPT-4 APIs, predicting whether the rationale supports the answer.
\begin{table*}[h]
\small
\centering
\begin{tabular}{l|l}
\toprule
system content & \parbox[c]{13cm}{%
\texttt{You are a helpful and precise assistant for following the given instruction.}
}\\
\midrule 
user content & \parbox[c]{13cm}{%
\texttt{[Instruction]\{Please read the question, rationale, and answer, and simply determine whether the answer can be derived from the rationale. Respond with ‘yes’ or ‘no’, without any explanations\}}\\ \\
\texttt{Task Description: \{Task Description\}} \\ \\
\texttt{Question: \{Question\}}\\ \\
\texttt{Rationale: \{Rationale\}}\\ \\
\texttt{Answer: \{Answer\}}\\
}\\
\bottomrule
\end{tabular}
\caption{Prompt template of evaluator for predicting whether the rationale supports the answer, given the question, rationale, and answer.}
\label{tab:evaluator-gen-CoTs}
\end{table*}

\begin{figure}[!h]
	\centering
	\includegraphics[width=0.90\linewidth]{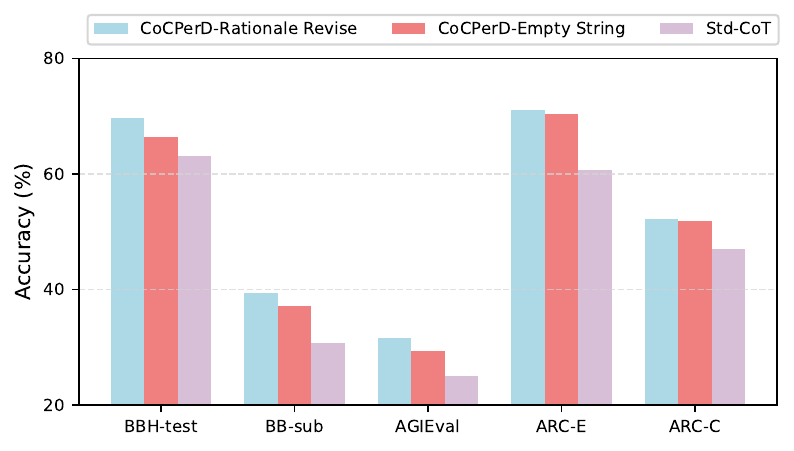}
    \vspace{-10pt}
	\centering
	\caption{Compare training CoPeD with different target outputs when the rationale is erroneous.}
	\label{ablation-on-task}
\end{figure}

\section{Analysis}
\label{outline-analysis}

\paragraph{What is the impact of training the student model with different target outputs when the rationale is erroneous?} We investigate the impact of training the student model to adopt different target outputs when the rationale is erroneous. As shown in Figure \ref{ablation-on-task}, the performance of the student model trained with an empty string as the target output when a reasoning error occurs is significantly lower than that of the student model trained with the correct rationale as the target. This suggests that the rationale correction task implicitly improves the quality of the rationales generated by the student model.
Furthermore, the performance of the student model trained with an empty string as the target output is notably superior to that of Std-CoT, which further demonstrates that CoPeD-TL enables the student model to benefit from the generated rationale when predicting answers, thereby effectively mitigating the spurious correlation between the question and the answer.

\paragraph{Whether the student model can effectively verify the correctness of the rationale?}

\begin{figure}[hbt]
	\centering
	\includegraphics[width=0.90\linewidth]{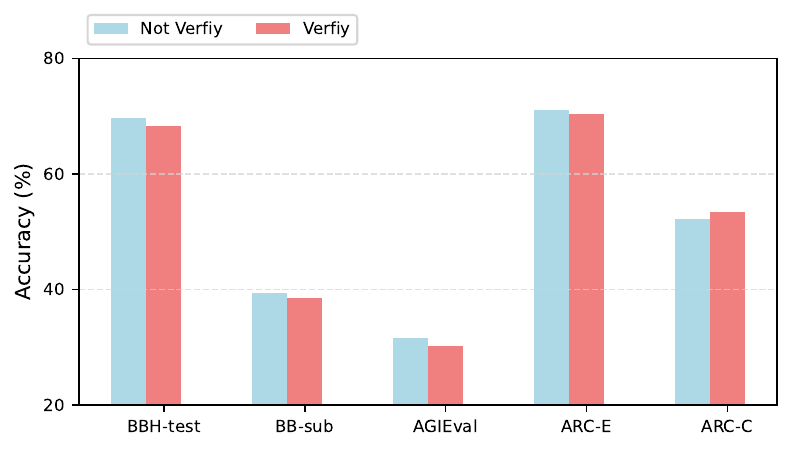}
    \vspace{-10pt}
	\centering
	% \caption{Comparison between CoPeD using the rationale status string ${rs}_{\mathrm{f}}$ as input and output. }
    \caption{Comparison between using the rationale status string ${rs}_{\mathrm{f}}$ as input and output in the correction task.}
	\label{ablation-on-plt-task-ve}
\end{figure}

We explore the impact of using the rationale status string ${rs}_{\mathrm{f}}$ as both input and output on CoPeD-TL’s performance in the rationale correction task on IND and OOD datasets. As shown in Figure \ref{ablation-on-plt-task-ve}, the experiment includes the following three settings: (1) input: When the rationale status string ${rs}_{\mathrm{f}}$ is used as input, the student model predicts the answer based on the generated rationale without verifying the correctness of the rationale; (2) output-correction: When the rationale status string ${rs}_{\mathrm{f}}$ is used as output, the student model, after identifying rationale errors, corrects the rationale and concatenates it with the question to re-predict the answer; (3) output-no correction: Even when the student model identifies rationale errors, the original rationale is used for prediction without any correction. The experimental results indicate that there is no significant performance difference between these three settings, suggesting that the student model is almost incapable of effectively verifying the correctness of the generated rationale. We believe the student model’s limited capacity, due to its smaller number of parameters, prevents it from independently verifying the correctness of the rationale, especially in complex reasoning tasks. Additionally, the model may struggle to generalize to different types of reasoning errors.
% particularly on out-of-distribution (OOD) datasets, where errors and data distributions differ from training, hindering accurate rationale verification.
% We believe that the limited capacity of the student model, due to its smaller number of parameters, leads to a lack of sufficient ability to independently verify the correctness of the rationale, especially when handling complex reasoning tasks. 

% We believe that the limited capacity of the student model, due to its smaller number of parameters, leads to a lack of sufficient ability to independently verify the correctness of the rationale, especially when handling complex reasoning tasks. Additionally, when faced with different types of reasoning errors, the student model may struggle to generalize across all scenarios, particularly on OOD  datasets, where the manifestation of reasoning errors and data distribution may differ from the training phase, preventing the model from accurately verifying the generated rationale.

\paragraph{Does the student model have the ability to correct erroneous rationale?}

\begin{figure}[hbt]
	\centering
	\includegraphics[width=0.90\linewidth]{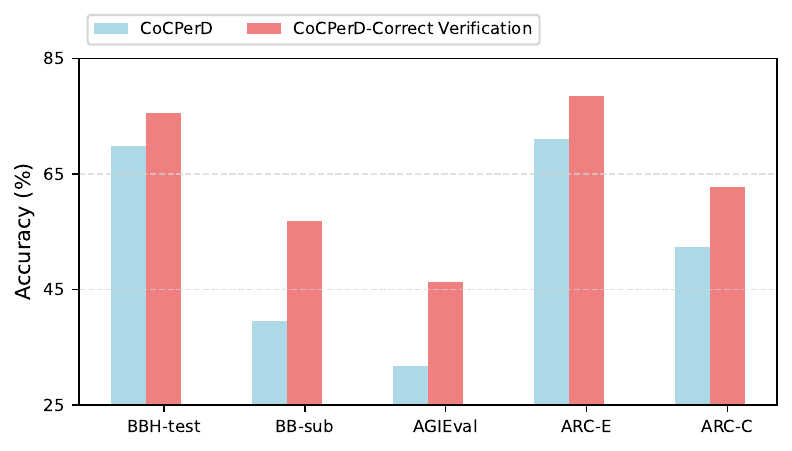}
    \vspace{-10pt}
	\centering
	\caption{Comparison between CoPeD without correcting rationale errors and correcting rationale errors under the assumption of correct rationale verification.}
	\label{ablation-on-plt-task-correct}
\end{figure}

We assume that the student model can correct verify erroneous rationales to evaluate its ability to correct them. During evaluation, the student model attempts to correct the rationales corresponding to previously erroneous answer predictions and then concatenates the corrected rationale with the question to re-predict the answer. As shown in Figure \ref{ablation-on-plt-task-correct}, The student model’s accuracy improves on both the IND and OOD datasets, mainly because 15\% to 30\% of the previous incorrect predictions are now correct. This suggests that the model can partially correct erroneous rationales, enhancing the final answer accuracy. Although the student model shows some limitations in correcting errors, this finding still reveals the substantial potential of distilling the ability to correct erroneous reasoning into student model.

\section{Case Study}
\label{appendix:case-study}

Table \ref{tab:bbh-web} shows that Std-CoT generates incorrect intermediate reasoning steps, leading to an incorrect final answer, indicating that Std-CoT struggles with effective reasoning in complex tasks. In contrast, CoPeD-TL generates a CoT that outperforms the teacher’s reasoning. Tables \ref{tab:reasoning_about_colored_objects} demonstrate that the intermediate reasoning steps generated by Std-CoT in domain-specific tasks lack causal relationships with the final answers, suggesting that Std-CoT causes the student model to capture spurious correlations between the question and the answer. In contrast, the CoT generated by CoPeD-TL effectively supports answer prediction, indicating an enhancement in the student’s reasoning ability. Tables \ref{tab:case-bb-winowhy} and \ref{tab:case-bb-implicit-relations} further show that in domain-adapted tasks, Std-CoT almost fails to reason correctly, whereas CoPeD-TL can clearly understand the question and provide logical and coherent reasoning.

\begin{table*}[!ht]
    \scriptsize
    \centering
    \renewcommand{\arraystretch}{1}
    \begin{tabular}{ll}
    \toprule
    \multicolumn{1}{l}{\textbf{Question:}} &
    \parbox[l]{13cm}{
Michael lies. Leda says Michael lies. Delbert says Leda tells the truth. Tamika says Delbert tells the truth. Fidel says Tamika lies. Does Fidel tell the truth?
    }\\
    \toprule
    \multicolumn{1}{l|}{\textbf{Teacher}}
    &
    \parbox[c]{13cm}{
    (1) Michael lies. So, we know that Michael lies.\\ \\ (2) Leda says Michael lies. Since we know from (1) that Michael lies, if Leda says Michael lies, then Leda tells the truth. \\\\(3) Delbert says Leda tells the truth. Since we know from (2) that Leda tells the truth, if Delbert says Leda tells the truth, then Delbert tells the truth. \\ \\ (4) Tamika says Delbert tells the truth. Since we know from (3) that Delbert tells the truth, if Tamika says Delbert tells the truth, then Tamika tells the truth. \\ \\ (5) Fidel says Tamika lies. Since we know from (4) that Tamika tells the truth, if Fidel says Tamika lies, then Fidel lies. \\ \\ Now, the question asks: Does Fidel tell the truth? We know from (5) that Fidel lies. \\ \\ Therefore, the answer is No.\quad \gtext{ \Large\textbf{\ding{51}}
    }}
    \\\midrule
    \multicolumn{1}{l|}{\textbf{Std-CoT}}
    &
    \parbox[c]{13cm}{
\rtext{According to this question, the answers to the questions are as follows: Michael lies: true. Leda lies: true. Delbert lies: true. Tamika lies: true. Fidel tells the truth: true.\\ \\ We know that Michael lies. Leda lies. Delbert lies. Tamika lies. Fidel tells the truth. \\ \\ To answer this question, we need to evaluate the statement Fidel tells the truth? \\ \\ Yes, Fidel tells the truth.} \\ \\ Therefore the answer is Yes. \quad\rtext{\Large\textbf{\ding{55}}}
    } 
    \\\midrule
    \multicolumn{1}{l|}{\textbf{CoPeD}}
    &
    \parbox[c]{13cm}{
    We first identify the parties in this dialogue as follows: \"Michael\" = M, \"Leda\" = L, \"Delbert\" = D, \"Tamika\" = T, and \"Fidel\" = F.\\ \\The statement \"Michael lies\" is expressed as M = lies. The statement \"Leda says Michael lies\" is expressed as L = lies. \\ \\ The statement \"Delbert says Leda tells the truth\" is expressed as D = tells-the-truth. \\ \\The statement \"Tamika says Delbert tells the truth\" is expressed as T = tells-the-truth. \\ \\ The statement \"Fidel says Tamika lies\" is expressed as F = lies. \\ \\Now, the question asks: Does Fidel tell the truth? We know from earlier that F = lies. \\ \\Therefore, the answer is No.\quad\gtext{\Large\textbf{\ding{51}}}
    } 
    \\
    \bottomrule
    \end{tabular}
    \caption{A web of lies case from BIG-Bench Hard.}
    \label{tab:bbh-web}
    \renewcommand{\arraystretch}{1}
\end{table*}

\begin{table*}[!ht]
    \scriptsize
    \centering
    \renewcommand{\arraystretch}{1}
    \begin{tabular}{ll}
    \toprule
    \multicolumn{1}{l}{\textbf{Question:}} &
    \parbox[l]{13cm}{
On the desk, you see a fuchsia dog leash and a teal necklace. Is the dog leash turquoise?\\Options: (A) yes (B) no
    }\\
    \toprule
    \multicolumn{1}{l|}{\textbf{Teacher}}
    &
    \parbox[c]{13cm}{
    According to this question, the color of the dog leash is fuchsia, not turquoise.\\\\Therefore, the answer is (B) no.\quad\gtext{\Large\textbf{\ding{51}}}
    } 
    \\\midrule
    \multicolumn{1}{l|}{\textbf{Std-CoT}}
    &
    \parbox[c]{13cm}{
    According to this question, the color of the dog leash is fuchsia. \\ \\ \rtext{Therefore, the answer is (A) yes.}\quad\rtext{\Large\textbf{\ding{55}}}
    } 
    \\\midrule
    \multicolumn{1}{l|}{\textbf{CoPeD}}
    &
    \parbox[c]{13cm}{
    According to this question, the color of the dog leash is fuchsia. \\\\ Therefore, the answer is (B) no.\quad\gtext{\Large\textbf{\ding{51}}}
    } 
    \\
    \bottomrule
    \end{tabular}
    \caption{A reasoning about colored objects case from BIG-Bench Hard.}
    \label{tab:reasoning_about_colored_objects}
    \renewcommand{\arraystretch}{1}
\end{table*}

\begin{table*}[!ht]
    \scriptsize
    \centering
    \renewcommand{\arraystretch}{1}
    \begin{tabular}{ll}
    \toprule
    \multicolumn{1}{l}{\textbf{Question:}} &
    \parbox[l]{13cm}{
Please answer the following questions about which words certain pronouns refer to. \\ The sun was covered by a thick cloud all morning, but luckily, by the time the picnic started, it was gone. The 'it' refers to the cloud because The cloud covered the sun. \\The above reasoning is\\Options: (A) correct (B) incorrect
    }\\
    \toprule
    \multicolumn{1}{l|}{\textbf{Teacher}}
    &
    \parbox[c]{13cm}{
First, we are given the information that the sun was covered by a thick cloud all morning.\\\\Then, we are told that by the time the picnic started, it was gone. The pronoun "it" is used to refer to something that was previously mentioned.\\\\In this case, the only thing mentioned that could be referred to by "it" is the cloud.\\\\Therefore, the answer is (A) correct.\quad\gtext{\Large\textbf{\ding{51}}}
    } 
    \\\midrule
    \multicolumn{1}{l|}{\textbf{Std-CoT}}
    &
    \parbox[c]{13cm}{
\rtext{The answer is not (A) because (A) states that the "it" refers to the picnic, which is not correct based on the context of the story.\\\\The "it" refers to the cloud because the cloud covered the sun.}\\\\Therefore, the answer is (B)\quad\rtext{\Large\textbf{\ding{55}}}
    } 
    \\\midrule
    \multicolumn{1}{l|}{\textbf{CoPeD}}
    &
    \parbox[c]{13cm}{
In this question, we are told that "The cloud covered the sun." A pronoun is used in the sentence to refer to the cloud.\\\\The sun is not directly mentioned in the sentence, but it is implied that the cloud covered the sun.\\\\Therefore, the cloud covered the sun.\\\\Therefore, the answer is (A) correct.\quad\gtext{\Large\textbf{\ding{51}}}
    } 
    \\
    \bottomrule
    \end{tabular}
    \caption{A winowhy case from BIG-Bench sub.}
    \label{tab:case-bb-winowhy}
    \renewcommand{\arraystretch}{1}
\end{table*}

\begin{table*}[!ht]
    \scriptsize
    \centering
    \renewcommand{\arraystretch}{1}
    \begin{tabular}{ll}
    \toprule
    \multicolumn{1}{l}{\textbf{Question:}} &
    \parbox[l]{13cm}{
Context and question: "But consider your daughters. Only think what an establishment it would be for one of them. Sir Adams and Lady Nelson are determined to go, merely on that account, for in general, you know, they visit no newcomers. Indeed you must go, for it will be impossible for us to visit him if you do not." "You are over-scrupulous, surely. I dare say Mr. Torres will be very glad to see you; and I will send a few lines by you to assure him of my hearty consent to his marrying whichever he chooses of the girls; though I must throw in a good word for my little Helga" "I desire you will do no such thing. Helga is not a bit better than the others; and I am sure she is not half so handsome as Olga, nor half so good-humoured as Freya. But you are always giving her the preference." What is Freya to Olga?\\Options: (A) Brother (B) Sister (C) Colleague (D) Master (E) Servant (F) Enemy (G) Lover (H) Husband (I) Father-in-law (J) Daughter-in-law (K) Mother-in-law (L) Son (M) Father (N) Wife (O) Grandfather (P) Grandson (Q) Grandmother (R) Son-in-law (S) Sister-in-law (T) Mother (U) Daughter (V) Cousin (W) Granduncle (X) Grandnephew (Y) Brother-in-law
    }\\
    \toprule
    \multicolumn{1}{l|}{\textbf{Teacher}}
    &
    \parbox[c]{13cm}{
First, we are given that Sir Adams and Lady Nelson are determined to go to visit someone. This suggests that the person they are visiting is important or has a high social status.\\\\Next, it is mentioned that Mr. Torres is the person they are visiting, and the speaker wants to send a message to him through the person they are addressing. The speaker also mentions that they give a good word for their "little Helga" but the person they are addressing disagrees and says that Helga is not better than the others.\\\\ \rtext{Based on this context, we can conclude that Freya, who is mentioned as being better-humored than Helga, is one of the daughters of the person being addressed.} \\\\Therefore, the answer is (U) Daughter.\quad\rtext{\Large\textbf{\ding{55}}}
    } 
    \\\midrule
    \multicolumn{1}{l|}{\textbf{Std-CoT}}
    &
    \parbox[c]{13cm}{
\rtext{the answer is not (B) or (C), as these are not the correct relationships.\\\\the answer is (D), because Freya is the daughter of Sir Adams and Lady Nelson, making her their daughter-in-law.} \\\\Therefore, the answer is D.\quad\rtext{\Large\textbf{\ding{55}}}
    } 
    \\\midrule
    \multicolumn{1}{l|}{\textbf{CoPeD}}
    &
    \parbox[c]{13cm}{
We have (1) "But consider your daughters. Only think what an establishment it would be for one of them."\\\\(2) "Sir Adams and Lady Nelson are determined to go, merely on that account, for in general, you know, they visit no newcomers." \\\\(3) "You are over-scrupulous, surely. I dare say Mr. Torres will be very glad to see you; and I will send a few lines by you to assure him of my hearty consent to his marrying whichever he chooses of the girls."\\\\(4) "You are always giving her the preference."\\\\From these clues, we can deduce that Freya is Olga's sister.\\\\Therefore, the answer is (B).\quad\gtext{\Large\textbf{\ding{51}}}
    } 
    \\
    \bottomrule
    \end{tabular}
    \caption{A implicit-relations case from BIG-Bench sub.}
    \label{tab:case-bb-implicit-relations}
    \renewcommand{\arraystretch}{1}
\end{table*}

\end{document}